# *W-Net*: One-Shot Arbitrary-Style Chinese Character Generation with Deep Neural Networks

Haochuan Jiang[1], Guanyu Yang[1], Kaizhu Huang[1(✉)], and Rui Zhang[2]

[1] Department of EEE, Xi'an Jiaotong - Liverpool University,
No. 111 Ren'ai Road, Suzhou, Jiangsu, People's Republic of China
`Kaizhu.huang@xjtlu.edu.cn`
[2] Department of MS, Xi'an Jiaotong - Liverpool University,
No. 111 Ren'ai Road, Suzhou, Jiangsu, People's Republic of China

**Abstract.** Due to the huge category number, the sophisticated combinations of various strokes and radicals, and the free writing or printing styles, generating Chinese characters with diverse styles is always considered as a difficult task. In this paper, an efficient and generalized deep framework, namely, the *W-Net*, is introduced for the one-shot arbitrary-style Chinese character generation task. Specifically, given a single character (one-shot) with a specific style (e.g., a printed font or hand-writing style), the proposed *W-Net* model is capable of learning and generating any arbitrary characters sharing the style similar to the given single character. Such appealing property was rarely seen in the literature. We have compared the proposed *W-Net* framework to many other competitive methods. Experimental results showed the proposed method is significantly superior in the one-shot setting.

## 1 Introduction

Chinese is a special language with both messaging functions and artistic values. On the other hand, Chinese contains thousands of different categories or over 10,000 different characters among which 3,755 characters, defined as level-1 characters, are commonly used. Given a limited number of Chinese characters or even one single character with a specific style (e.g., a personalized hand-writing calligraphy or a stylistic printing font), it is interesting to mimic automatically many other characters with the same specific style. This topic is very difficult and rarely studied simply because of the large category number of different Chinese characters with various styles. This problem is even harder due to the unique nature of Chinese characters among which each is a combination of various strokes and radicals with diverse interactive structures.

Despite these challenges, there are recently a few proposals relevant to the above-mentioned generation task. For example, in [13], strokes are represented by time-series writing evenly-thick trajectories. Then it is sent to the Recurrent Neural Network based generator. In [6], font feature reconstruction for standardized character extraction is achieved based on an additional network to assist the one-to-one image-to-image translation framework. Over 700 pre-selected training images are needed in this framework. In the *Zi2Zi* [12] model, a one-to-many mapping is achieved with only a single model by the fixed Gaussian-noise based categorical embedding with over 2,000 training examples per style.

There are several main limitations in the above approaches. On one hand, the performance of these methods usually relies heavily on a large number of samples with a specific style. In the case of a few-shot or even one-shot generation, these methods would fail to work. On the other hand, these methods may not be able to transfer to a new style which has not been seen during training. Such drawbacks may hence present them from being used practically.

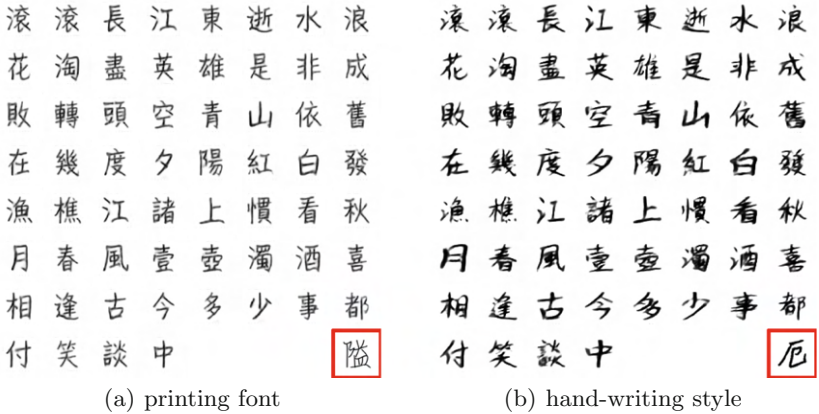

(a) printing font            (b) hand-writing style

**Fig. 1.** Generated traditional characters by the proposed *W-Net* model **with one single sample available** (the right-bottom character with red boxes). (Color figure online)

In this paper, aiming to generate Chinese characters when even given one shot sample with a specific arbitrary style (seen or unseen in training), we propose a novel deep model named *W-Net* as a generalized style transformation framework. This framework better solves the above-mentioned drawbacks and could be easily used in practice. Particularly, inherent from the *U-Net* framework [9] for the one-to-one image-to-image translation task [4], the proposed *W-Net* employs two parallel convolution-based encoders to extract style and content information respectively. The generated image will be obtained by the deconvolution-based decoder by using the encoded information. Short-cut connections [9] and multiple residual blocks [2] are set to deal with the gradient vanishing problem and balance information from both encoders to the decoder. The training of the *W-Net* follows an adversarial manner. Inspired by the recently proposed Wasserstein Generative Adversarial Network (W-GAN) framework with gradient penalty [1], an independent discriminator[1] ($D$) are employed to assist the *W-Net* ($G$) learning.

---
[1] The discriminator actually attaches an auxiliary classifier proposed in [8].

As a methodological guidance, only one-shot arbitrary-style Chinese character generation is demonstrated in this paper, as examples given in Fig. 1. However, the *W-Net* framework can be extended to a variety of related topics on one-shot arbitrary-style image generation. With such a proposal, the data synthesizing tasks with few samples available can be fulfilled much more readily and effectively than previous approaches in the literature.

## 2  Model Definition

### 2.1  Preliminary

Denote $X$ be a Chinese character dataset, consisting of $J$ different characters with in total $I$ different fonts. Let $x_j^i$ be a specific sample in $X$, regarded as the **real target**. Following [3,5], the superscript $i \in [0, 1, 2, ..., I]$ represents $i$-th style, while the subscript $j \in [1, 2, ..., J]$ denotes $j$-th example.

Specifically, during the training, when $i = 0$, $x_j^0$ denotes the image of the $j$-th character with a standardized style information, named as the **prototype content**. Meanwhile, $x_k^i, k \in [1, 2, ..., J]$ is defined as a **style reference** equipping with the $i$-th style information, the same as $x_j^i$. Be noted that commonly, $j$ and $k$ are different. In the proposed model, each $x_j^i$ is assumed to be combined with information from the prototype $x_j^0$ and the $i$-th writing style learned from $x_k^i$. The proposed *W-Net* model will then produce the **generated target** $G(x_j^0, x_k^i)$ which is similar to $x_j^i$ by taking both $x_j^0$ and $x_k^i$ simultaneously.

Be noted that only single style character is required to produce the generated target. It is defined as the **One-Shot Arbitrary-Style Character Generation** task. Specifically, the given single sample (E.g., $x_p^m$, where $m$ can be any arbitrary style, meanwhile $p$ could be any single character. Both $m$ and $p$ can be irrelevant to $[1, 2, ..., I]$ and $[1, 2, ..., J]$ respectively) is seen as the **one-shot** style reference. The task can be readily fulfilled by feeding any content prototype $(x_q^0)$ of the desired $q$-th character on condition of those relevant outputs of the $Enc_r$ (to be connected to the $Dec$ with both shortcut or residual block connections, as will be demonstrated in Sect. 2.2) to produce $G(x_q^0, x_p^m)$ given the single style example $(x_p^m)$. In such the setting, alternating $q$ will lead to synthesizing different characters. Simultaneously, all the generated examples are expected to imitate the $m$-th style information given by $x_p^m$. Similarly, $q$ could also be out of $[1, 2, ..., J]$ as well.

### 2.2  *W-Net* Architecture

Figure 2 illustrates the basic structure of the proposed *W-Net* model. It consists of the content prototype encoder ($Enc_p$, the blue part), the style reference encoder ($Enc_r$, the green part), and the decoder ($Dec$, the red part).

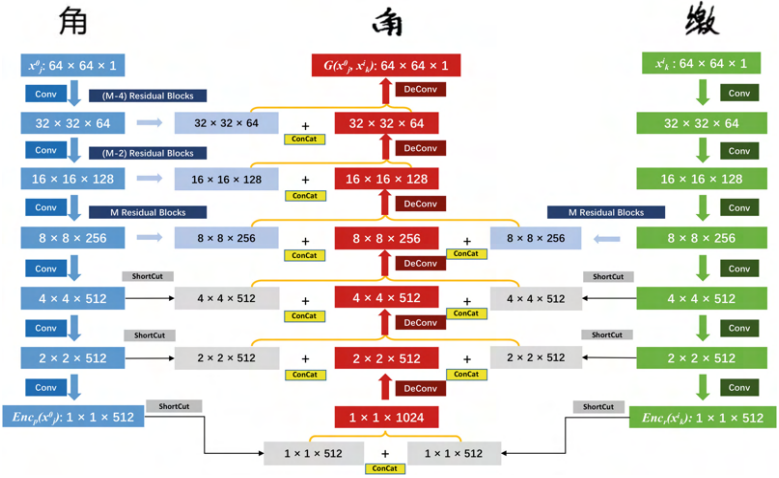

**Fig. 2.** The *W-Net* (better viewed in colors), where the blue part represents $Enc_p$, green for $Enc_r$, and red for $Dec$. Conv: $5 \times 5$ convolution; DeConv: $5 \times 5$ deconvolution. Fixed stride 2 and ReLU are applied to both Conv and DeConv. ConCat: Feature concatenation on the channel; ShortCut: Feature Shortcut.

The $Enc_p$ and $Enc_r$ are constructed as sequences of convolutional layers, where $5 \times 5$ filters with fixed stride 2 and ReLU function are implemented. By this setting, $64 \times 64$ prototype and reference images $x_j^0$ and $x_j^k$ will be mapped into $1 \times 512$ feature vector, denoted as $Enc_p(x_j^0)$ and $Enc_r(x_j^k)$ respectively.

Identical to the decoder in the *U-Net* framework [9], *Dec* is designed as a deconvolutional progress layer-wisely connected with $Enc_p$ and $Enc_r$. It produces a generated image, the size of which is consistent with all the input images of both encoders. Specifically, for higher-level features between the decoder and both the encoders, connections are achieved by simple feature shortcut. For lower-level layers of $Enc_p$, a series of residual blocks[2] [2] are applied and connected to the *Dec*. The number of blocks is controlled by a super parameter $M$. On the contrast, as the writing style is a kind of high-level deep feature, there is only one residual block connection (with $M$ blocks) between $Enc_r$ and *Dec*, omitting lower-level feature concatenation at the same time.

### 2.3  Optimization Strategy and Losses

The proposed *W-Net* is trained adversarially based on the Wasserstein Generative Adversarial Network (W-GAN) framework, regarded as the generator $G$. Specifically, it takes the content prototype and the style reference, and then returns generated target as $G(x_j^0, x_k^i) = Dec(Enc_p(x_j^0), Enc_r(x_k^i))$ closed to $x_j^i$. $G$ is optimized by taking advantages of the adversarial network $D$ as well as several optimization losses defined as follows.

---

[2] The structure of the residual block follows the setting in [6].

**Training Strategy:** The learning of the *W-Net* follows the adversarial training scheme. In each learning iteration, there are two independent procedures, including the $G$ training and the $D$ training respectively. The $G$ and the $D$ are trained to optimize Eqs. (1) and (2) respectively.

$$\mathbb{L}_G = -\alpha \mathbb{L}_{adv-G} + \beta_d \mathbb{L}_{dac} + \beta_p \mathbb{L}_{enc-p-cls} + \beta_r \mathbb{L}_{enc-r-cls} \\ + \lambda_{l1}\mathbb{L}_1 + \lambda_\phi \mathbb{L}_\phi + \psi_p \mathbb{L}_{Const_p} + +\psi_r \mathbb{L}_{Const_r} \quad (1)$$

$$\mathbb{L}_D = \alpha \mathbb{L}_{adv-D} + \alpha_{GP} \mathbb{L}_{adv-GP} + \beta_d \mathbb{L}_{dac} + \beta_p \mathbb{L}_{enc-p-cls} + \beta_r \mathbb{L}_{enc-r-cls} \quad (2)$$

**Adversarial Loss:** $G$ optimizes $\mathbb{L}_{adv-G} = D(x_j^0, G(x_j^0, x_k^i), x_k^i)$, while $D$ minimizes $\mathbb{L}_{adv-D} = D(x_j^0, x_j^i, x_k^i) - D(x_j^0, G(x_j^0, x_k^i), x_k^i)$. Be noted that a gradient penalty is set as $\mathbb{L}_{adv-GP} = ||\nabla_{\widehat{x}} D(x_j^0, \widehat{x}, x_k^i) - 1||_2$ [1], where $\widehat{x}$ is uniformly interpolated along the line between $x_j^i$ and $G(x_j^0, x_k^i)$.

**Categorical Loss of the Discriminator Auxiliary Classifier:** $\mathbb{L}_{dac} = \left[\log C_{dac}(i|x_j^0, x_j^i, x_k^i)\right] + \left[\log C_{dac}(i|x_j^0, G(x_j^0, x_k^i), x_k^i)\right]$, inspired by [8].

**Reconstruction Losses** consists the pixel-level difference ($\mathbb{L}_1 = ||(x_j^i - G(x_j^0, x_k^i))||_1$) and the high-level feature variation ($\mathbb{L}_\phi = \sqrt{\sum_\phi \left[\phi(x_j^i) - \phi(G(x_j^0, x_k^i))\right]^2}$). $\phi(.)$ represents a specific deep feature. The VGG-16 network [10] trained with multiple character styles is employed here. In this optimization, in total five convolutional features including $\phi_{1-2}$, $\phi_{2-2}$, $\phi_{3-3}$, $\phi_{4-3}$, $\phi_{5-3}$ are involved.

**Constant Losses of the Encoders:** The constant losses [11] are also employed for both encoders. They are given by $\mathbb{L}_{Const_p} = ||Enc_p(x_j^0) - Enc_p(G(x_j^0, x_k^i))||^2$ and $\mathbb{L}_{Const_r} = ||Enc_r(x_k^i) - Enc_r(G(x_j^0, x_k^i))||^2$ respectively for $Enc_p$ and $Enc_r$.

**Categorical Losses on Both Encoders:** To ensure the specific functionalities of the two encoders, we forced the content and style features extracted by them to be equipped with the corresponding commonality separately for the same kind. It is designated by adding a fully-connecting to implement the category classification task, which leads to that both encoders learn their own representative features, simultaneously over-fitting is avoided thereby. $\theta_p$ and $\theta_r$ are used to denote the fully-connecting and softmax functions together for both output feature vectors of encoders respectively, while the classifications are noted as $C_{encp}$ and $C_{encr}$. The upon-mentioned cross entropy losses of both classifications are given as $\mathbb{L}_{enc-p-cls} = \left[\log C_{encp}(j|\theta_p(Enc_p(x_j^0))\right]$ and $\mathbb{L}_{enc-p-cls} = \left[\log C_{encr}(i|\theta_r(Enc_r(x_k^i))\right]$ respectively. Be noted that $i$ and $j$ represent the specific style and the character labels.

## 3  Experiment

A series of experiments have been conducted to verify the effectiveness of the proposed *W-Net* network. Both printed and hand-writing fonts are evaluated. Several relevant baselines are also referred for the comparison as well.

### 3.1  Experiment Setting

80 fonts are specially chosen in standard Chinese printed font database. 50 of them, each containing 3,755 level-1 simplified Chinese characters, are involved in the training set. The offline version of both CASIA-HWDB-1.1 (for simplified isolated characters) and the CASIA-HWDB-2.1 (for simplified cursive characters) [7] are involved as the hand-writing data set. Characters written by 50 writers (No. 1,101 to 1,150) are selected as the training set, resulting in total 249,066 samples (4,980 examples per writer averagely). For both sets, the testing data are chosen due to different evaluation purposes. *HeiTi* (boldface font) is used as the prototype font for both the sets, as examples given in Fig. 3(a).

Baseline models include two upgraded version of the *Zi2Zi* [12] framework which were modified for the few-shot new-coming style synthesization task. One utilizes a fine-tuning strategy (noted as *Zi2Zi-V1*), where the style information is assumed to be the linear combination of multiple known styles represented by the fixed Gaussian-noise based categorical embedding; The other (*Zi2Zi-V2*) discards the categorical embedding by introducing the final softmax output of a pre-trained VGG-16 network (embedder network), identical to the one employed in Sect. 2.3. All the other network architecture and training settings of these baselines are all the same by following [12]. Characters from both databases are represented by $64 \times 64$ gray-scale images, after which they are then binarized. One thing to be particularly noted is that both the proposed *W-Net* and the *Zi2Zi-V2* follow the **one-shot** setting, where only single style example ($x_p^m$) is referred during the evaluation process. However, the *Zi2Zi-V1* employ the few-shot (32 references) scheme in order to obtain a valid fine-tuning performance.

Hyper-parameters during *W-Net* training are tuned empirically and set as follows: residual block number is $M = 5$; relevant penalties are: $\alpha = 3$, $\alpha_{GP} = 10$, $\beta_d = 1$, $\beta_p = \beta_r = 0.2$, $\lambda_{l1} = 50$, $\lambda_\phi = 75$, $\psi_p = 3$ and $\psi_r = 5$.

The Adam optimizer with $\beta_1 = 0.5$ and $\beta_2 = 0.999$ is implemented, while the initial learning rate is set to be 0.0005 and decayed exponentially after each training epoch. The architecture of $D$ follows the setting of the *Zi2Zi* framework [12] with the W-GAN framework. For the sake of speeding up and stabilizing the training progress, the batch normalization is applied several layers to the $G$ network, while the layer normalization is selected for $D$. Drop out trick is also applied to both $G$ and $D$ to improve the generalization performance. Weight decay is also engaged to avoid the over-fitting issue. The proposed *W-Net* framework and other baselines are implemented with the Tensorflow (r1.5).

### 3.2  Model Reasonableness Evaluation

The *W-Net* model is verified by setting $p = q$ for content $x_q^0$ and the style $x_p^m$ in this section. Hereby, the reference is exactly the real target ($x_p^m = x_q^m$). For each evaluation, as previously instructed, only single style reference ($x_p^m$, characters of 2nd rows in (b)–(e) of Fig. 3) is engaged. The generated image is seen to follow the style tendency of the one-shot reference if the proposed *W-Net* is capable of reconstructing the extracted style information in the reference image $x_p^m$.

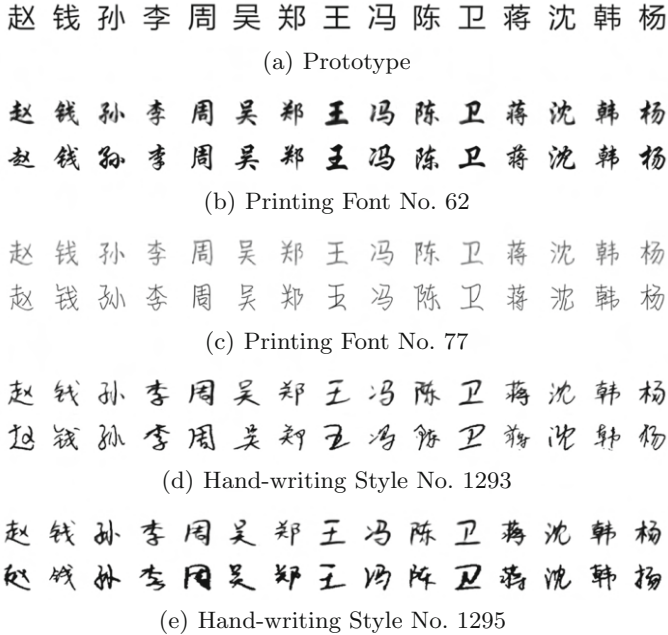

**Fig. 3.** Several examples of generated data of unseen printing and hand-writing styles. (a) The input content prototypes; (b)–(e) 1st row: generated characters; 2nd row: corresponding style references (ground truth characters).

Figure 3 illustrates several examples of the comparison result for synthesizing unseen styles during training. It can be observed that styles of both printed and hand-writing types are learned and transferred to the prototypes by the *W-Net* model with the proper performance by maintaining the style consistency.

### 3.3  Model Effectiveness Evaluation

The effectiveness of *W-Net* is tested by generating commonly-used Chinese characters (simplified and traditional) with alternative styles. In this setting, $x_p^m$ are randomly selected one-shot character with the $m$-th style information to imitate the real application scenario, while $q$ are referred to the desired content prototypes to be generated. Commonly, $p \neq q$.

Figures 4 and 5 list several examples of the generated images by *W-Net* and two baselines for seen and unseen styles during training respectively. Particularly, only simplified Chinese characters are accessible during training, as seen in the left four columns of each subfigure of those two illustrations. There is no ground truth data in both the involved databases of those traditional images in other remaining columns.

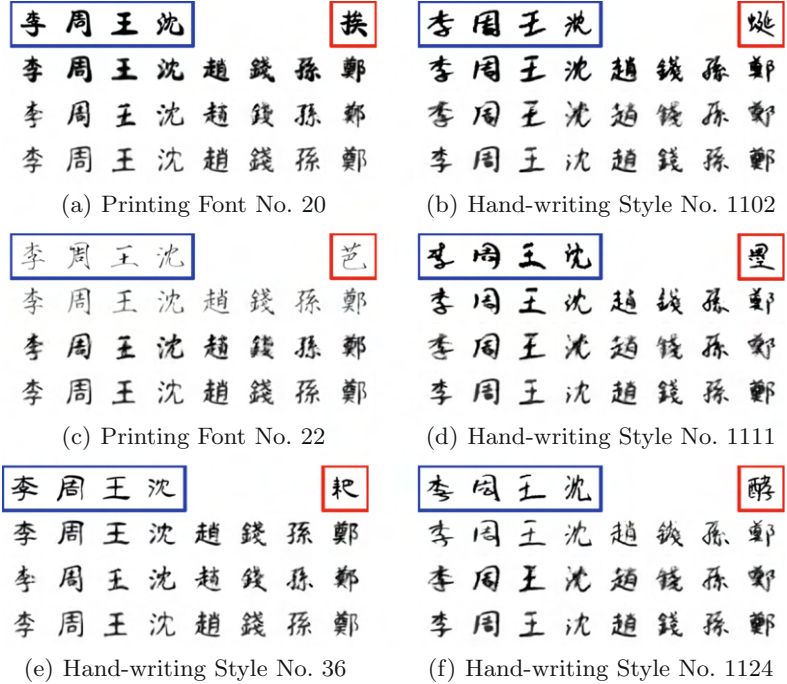

**Fig. 4.** Several examples of generated characters of seen styles. (a), (c) and (e) are printing fonts; (b), (d) and (f) are hand-writing styles. In each figure, 1st row: ground truth characters (with blue boxes) and the one-shot style reference (with red boxes); 2nd: *W-Net* generated characters; 3rd row: *Zi2Zi-V1* performance; 4th row: *Zi2Zi-V2* performance. (Color figure online)

When generating characters with a specific seen style during training, it can be intuitively observed in Fig. 4 that even given one-shot style reference, the generated fonts by *W-Net* look very similar to the corresponding real targets. Differently, under the few-shot setting, the *Zi2Zi-V1* still produces blurred images, while *Zi2Zi-V2* seems to synthesize characters with the averaged style. The proposed *W-Net* outperforms others by producing characters with both desired contents and consistent styles with only one-shot style reference available.

Simultaneously, acceptable generations can still be obtained from the Fig. 5 by the proposed scheme when constructing unseen styles with one-shot style reference as well. Though the generated samples are not similar enough as that in previous examples, a clear stylistic tendency can still be clearly observed. On the contrast, *Zi2Zi-V1* failed to produce high-quality images even 32 references are given for the fine-tuning due to the over-fitting issue. At the same time, the *Zi2Zi-V2* failed to generate distinguishable styles since it is only capable of learning styles from the original basis provided by the embedder network (VGG-16).

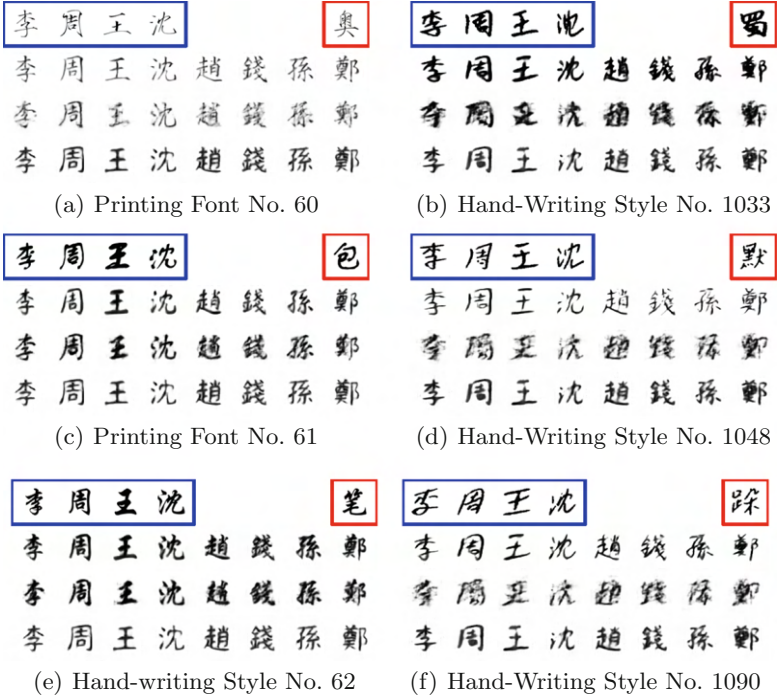

**Fig. 5.** Several examples of generated characters of unseen styles. (a), (c) and (e) are printing fonts; (b), (d) and (f) are hand-writing styles. In each figure, 1st row: ground truth characters (with blue boxes) and the one-shot style reference (with red boxes); 2nd: *W-Net* generated characters; 3rd row: *Zi2Zi-V1* performance; 4th row: *Zi2Zi-V2* performance. (Color figure online)

### 3.4  Analysis on Failure Examples

The proposed model would sometimes fail to capture the style information when it is over far away from the prototype font. For example, some cursive writing may play a negative role in the generation process since input contents are all isolated characters. Some failed generated characters are given in Fig. 6, of which the 2nd row lists the corresponding one-shot style references.

Upon the proposed *W-Net*, each target is regarded as a non-linear style transformation from a reference to a prototype. However, when the style is too different from the content font, the model fails to learn this complicated mapping relationship. In such extreme circumstances, the provided single prototype font in this paper might be an inappropriate choice. In this case, it could be a good idea to learn additional mappings which can transform the original prototype to suitable latent features so as to better handle free writing styles in real scenarios.

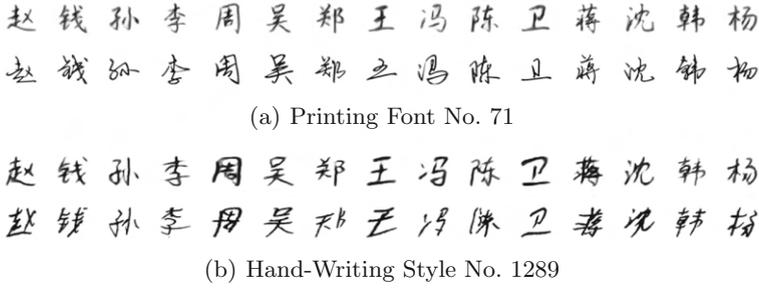

(a) Printing Font No. 71

(b) Hand-Writing Style No. 1289

**Fig. 6.** Unsatisfied generated examples. In each figure: 1st row: generated characters; 2nd row: corresponding style references (ground truth characters)

## 4 Conclusion and Future Work

A novel generalized framework named *W-Net* is introduced in this paper in order to achieve one-shot arbitrary-style Chinese character generation task. Specifically, the proposed model, composing of two encoders and one decoder with several layer-wised connections, is trained adversarially based on the W-GAN scheme. It enables synthesizing any arbitrary stylistic character by transferring the learned style information from one single reference to the input content prototype. Extensive experiments have demonstrated the reasonableness and effectiveness of the proposed *W-Net* model in the one-shot setting.

Extensions to more proper mapping architectures for image reconstruction will be studied in the future so as to capture sufficiently complicated and free writing styles. Meanwhile, practical applications are to be developed not only restricted in the character generation domain, but also in other relevant arbitrary-style image generation tasks.

**Acknowledgements.** The work was partially supported by the following: National Natural Science Foundation of China under no. 61473236 and 61876155; Natural Science Fund for Colleges and Universities in Jiangsu Province under no. 17KJD520010; Suzhou Science and Technology Program under no. SYG201712, SZS201613; Jiangsu University Natural Science Research Programme under grant no. 17KJB-520041; Key Program Special Fund in XJTLU under no. KSF-A-01 and KSF-P-02.